\documentclass[11pt]{article} % For LaTeX2e

\usepackage{rldmsubmit,palatino}
\usepackage{graphicx}
\usepackage{hyperref}
\usepackage{subfig}
\usepackage[utf8]{inputenc}
\hypersetup{colorlinks=true,urlcolor=[rgb]{0, 0, 1}, citecolor=.}
\usepackage[
  separate-uncertainty = true,
  multi-part-units = repeat
]{siunitx}
\usepackage{amsmath,amssymb}
\DeclareMathOperator{\E}{\mathbb{E}}

\title{Hacking Google reCAPTCHA v3 using Reinforcement Learning}

\author{
Ismail Akrout\thanks{equal contribution} \\
Télécom ParisTech\\
\texttt{akrout.ismail@gmail.com} \\
\And
Amal Feriani\footnotemark[1] \\
Ankor AI\\
\texttt{amal.feriani@gmail.com} \\
\AND
Mohamed Akrout\\
University of Toronto\\
\texttt{makrout@cs.toronto.edu} \\
}

\begin{document}

\maketitle

\begin{abstract}
%The \emph{title} should be a maximum of 100 characters. 
%The \emph{abstract} should be a maximum of 2000 characters of text, including spaces (no figure is allowed). You will be asked to copy this into a text-only box; and it will appear as such in the conference booklet. Use 11~point type, with a vertical spacing of 12~points.  The word \textbf{Abstract} must be centered, bold, and in point size 12. Two line spaces precede the abstract.\\\\
We present a Reinforcement Learning (RL) methodology to bypass Google reCAPTCHA v3. We formulate the problem as a grid world where the agent learns how to move the mouse and click on the reCAPTCHA  button to receive a high score. We study the performance of the agent when we vary the cell size of the grid world and show that the performance drops when the agent takes big steps toward the goal. Finally, we use a divide and conquer strategy to defeat the reCAPTCHA system for any grid resolution. Our proposed method achieves a success rate of $97.4\%$ on a $100\times100$ grid and $96.7\%$ on a $1000\times1000$ screen resolution.

%prior to the QA model and the image textual description, we greedily ask the best symptom that maximizes the information gain over symptoms. We demonstrate that combining the QA model with the CNN increases the accuracy up to 10\% as compared to the CNN alone, and more than 30\% as compared to the QA model alone.
\end{abstract}

\keywords{Reinforcement Learning, reCAPTCHA, Security, Artificial Intelligence, Machine Learning}

\acknowledgements{We thank Douglas Tweed for his valuable feedback and helpful discussions.}

\startmain % to start the main 1-4 pages of the submission.
\section{Introduction}
Artificial Intelligence (AI) has been experiencing unprecedented success in the recent years thanks to the progress accomplished in Machine Learning (ML), and more specifically Deep Learning (DL). These advances raise several questions about AI safety and ethics \cite{Amodei2016}. In this work, we do not provide an answer to these questions but we show that AI systems based on ML algorithms such as reCAPTCHA v3 \cite{Captchav3} are still vulnerable to automated attacks.
Google's reCAPTCHA system, for detecting bots from humans, is the most used defense mechanism in websites. Its purpose is to protect against automated agents and bots, attacks and spams. Previous versions of Google's reCAPTCHA (v1 and v2) present tasks (images, letters, audio) easily solved by humans but challenging for computers. The reCAPTCHA v1 presented a distorted text that the user had to type correctly to pass the test. This version was defeated by Bursztein  et al. \cite{Bursztein2014} with 98\% accuracy using ML-based system to segment and recognize the text. As a result, image-based and audio-based reCAPTCHAs were introduced as a second version. Researchers have also succeeded in hacking these versions using ML and more specifically DL. For example, the authors in \cite{Bock2017} designed an AI-based system called \textit{UnCAPTCHA} to break Google's most challenging audio reCAPTCHAs. On 29 October 2018, the official third version was published \cite{Captchav3-announcement} and removed any user interface. Google's reCAPTCHA v3 uses ML to return a risk assessment score between 0.0 and 1.0. This score characterize the trustability of the user. A score close to 1.0 means that the user is human.\\

In this work, we introduce an RL formulation to solve this reCAPTCHA  version. Our approach is programmatic: first, we propose a plausible formalization of the problem as a Markov Decision Process (MDP) solvable by state-of-the-art RL algorithms; then, we introduce a new environment for interacting with the reCAPTCHA  system; finally, we analyze how the RL agent learns or fails to defeat Google reCAPTCHA. Experiment results show that the RL agent passes the reCAPTCHA test with $97.4$ accuracy. To our knowledge, this is the first attempt to defeat the reCAPTCHA v3 using RL . 

\iffalse
In summary, this paper makes the following distinct contributions:
\begin{itemize}
    \item We show how to formulate the user's mouse movement as a learning task in a RL environment;
    \item We present a RL agent capable of defeating the newest version of reCAPTCHA;
    \item We develop an environment to simulate the user experience with websites using the reCAPTCHA system;
    \item We propose a scalable and efficient method to defeat reCAPTCHA on different environment's sizes.
\end{itemize}
\fi

\section{Method}
\subsection{Preliminaries}
An agent interacting with an environment is modeled as a Markov Decision Process (MDP) \cite{puterman2014markov}. A MDP is defined as a tuple $(\mathcal{S}, \mathcal{A}, P, r)$ where $\mathcal{S}$ and $\mathcal{A}$ are the sets of possible states and actions respectively. $P(s,a,s^{'})$ is the transition probabilities between states and $r$ is the reward function. Our objective is to find an optimal policy $\pi^*$ that maximizes the future expected rewards. Policy-based methods directly learn $\pi^{*}$. Let's assume that the policy is parameterized by a set of weights $w$ such as $\pi=\pi(s,w)$. Then, the objective is defined as: $ J(w) = \E_{\pi}\bigg[\sum_{t=0}^T \gamma^t r_t\bigg] $
where $\gamma$ is the discount factor and $r_t$ is the reward at time $t$.

Thanks to the policy gradient theorem and the gradient trick \cite{sutton2018reinforcement}, the Reinforce algorithm \cite{Williams1992} estimates gradients using (\ref{eq:4}).
\begin{equation} \label{eq:4}
\nabla \mathbb{E}_{\pi}\bigg[\sum_{t=0}^T \gamma^t r_t\bigg] = \mathbb{E}_{\pi} \bigg[\sum_{t=0}^T \nabla \log \pi(a_t|s_t)R_t\bigg]
\end{equation}
$R_t$ is the future discounted return at time $t$ defined as $R_t = \sum_{k=t}^T \gamma^{(k-t)}\cdot r_k$, where $T$ marks the end of an episode.

Usually the equation (\ref{eq:4}) is formulated as the gradient of a loss function $L(w)$ defined as follows: $L(w) = - \frac{1}{N} \sum_{i=1}^{N} \sum_{t=0}^T \nabla \log \pi(a^{i}_t|s^{i}_t)R^i_t $ where $N$ is the a number of collected episodes.

\subsection{Settings}
To pass the reCAPTCHA  test, a human user will move his mouse starting from an initial position, perform a sequence of steps until reaching the reCAPTCHA  check-box and clicking on it. Depending on this interaction, the reCAPTCHA  system will reward the user with a score. In this work, we modeled this process as a MDP where the state space $\mathcal{S}$ is the possible mouse positions on the web page and the action space is $\mathcal{A}=\{up, left, right, down\}$. Using these settings, the task becomes similar to a grid world problem.

%%%
\begin{figure}[h]
\centering
\frame{\includegraphics[width=0.35\linewidth]{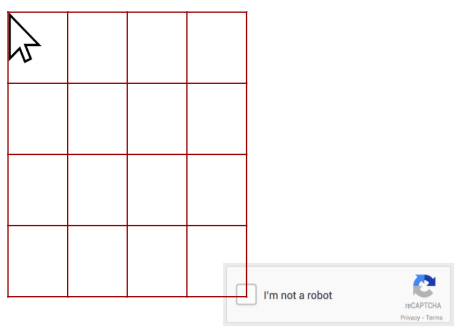}}
\caption{The agent's mouse movement in a MDP}
\label{fig:captcha-env}
\end{figure}

As shown in Figure \ref{fig:captcha-env}, the starting point is the initial mouse position and the goal is the position of the reCAPTCHA  is the web page. For each episode, the starting point is randomly chosen from a top right or a top left region representing 2.5\% of the browser window's area (5\% on the x-Axis and 5\% on the y-Axis). A grid is then constructed where each pixel between the initial and final points is a possible position for the mouse. We assume that a normal user will not necessary move the mouse pixel by pixel. Therefore, we defined a cell size $c$ which is the number of pixels between two consecutive positions. For example, if the agent is at the position $(x_0,y_0)$ and takes the action $left$, the next position is then $(x_0-c,y)$.

One of our technical contributions consists in our ability to simulate the same user experience as any normal reCAPTCHA  user. This was challenging since reCAPTCHA  system uses different methods to distinguish fake or headless browsers, inorganic behaviors of the mouse, etc. Our environment overcomes all these problems. For more details about the environment implementation, refer to section~\ref{sec:details-implementation}. At each episode, a browser page will open up with the user mouse at a random position, the agent will take a sequence of actions until reaching the reCAPTCHA  or the horizon limit $T$ defined as twice the grid diagonal i.e. $T = 2 \times \sqrt{a^2 + b^2}$ where $a$ and $b$ are the grid's height and width respectively. Once the episode ends, the user will receive the feedback of the reCAPTCHA  algorithm as would any normal user. 

\section{Experiments and Results}
We trained a Reinforce agent on a grid world of a specific size. Our approach simply applies the trained policy to choose optimal actions in the reCAPTCHA  environment. Our results presented are the success rates across $1000$ runs. We consider that the agent successfully defeated the reCAPTCHA  if it obtained a score of $0.9$. In our experiments, the discount factor was $\gamma=0.99$. The policy network was a vanilla two fully connected layer network. The parameters were learned with a learning rate of $10^{-3}$ and a batch size of $2000$. Figure \ref{fig:dc-captcha-results} shows the results for a $100\times100$ grid. Our method successfully passed the reCAPTCHA  test with a success rate of $97.4\%$.% These numbers indicate that even if the agent was trained with a different reward function, he/she is able to to beat the CAPTCHA with high scores.
%\setlength{\tabcolsep}{1em} % for the horizontal padding
%\begin{table}[h]
%\centering
%\caption{The distribution of rewards for a $100\times100$ grid}
%\label{tab:success-rate}
%\renewcommand{\arraystretch}{1.4}% for the vertical padding

%\begin{tabular}{c|c|c|c|c|c|}
%\cline{2-6}
                                          %& \multicolumn{5}{c|}{\textbf{reCAPTCHA v3 Reward}}                                   \\ \cline{2-6} 
                                          %& \textbf{0} & \textbf{0.1} & \textbf{0.3} & \textbf{0.7} & \textbf{0.9} \\ \hline
%\multicolumn{1}{|c|}{\textbf{Percentage}} &       0.1     &          2.4    &      0.1        &      0.1        &     97.4         \\ \hline
%\end{tabular}
%\end{table}

Next, we consider testing our method on bigger grid sizes. If we increase the size of the grid, the state space dimension $|\mathcal{S}|$ increases exponentially and it is not feasible to train a Reinforce algorithm with a very high dimensional state space. For example, if we set the grid size to $1000\times1000$ pixels, the state space becomes $10^6$ versus $10^4$ for a $100\times100$. This is another challenge that we address in this paper: how to attack the reCAPTCHA  system for different resolutions without training an agent for each resolution?

\section{An efficient solution to any grid size}
\begin{figure}[h]
\centering
\includegraphics[width=0.3\linewidth]{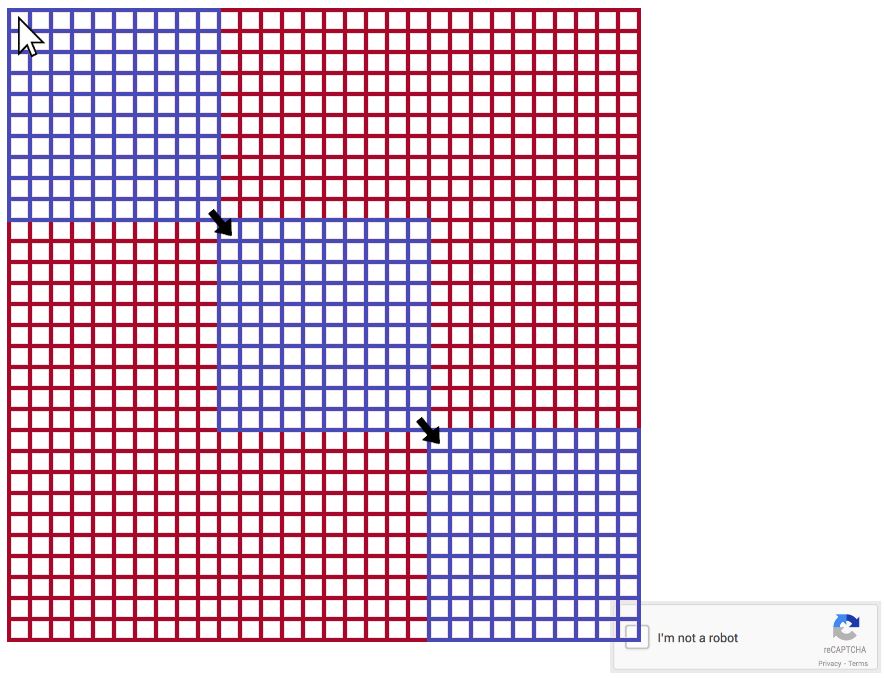}
\caption{Illustration of the divide and conquer approach: the agent  runs sequentially on the diagonal grid worlds in purple. The grid worlds in red are not explored.}
\label{fig:dc-captcha}
\end{figure}

\begin{figure}[h!]
\centering
\includegraphics[width=0.55\linewidth]{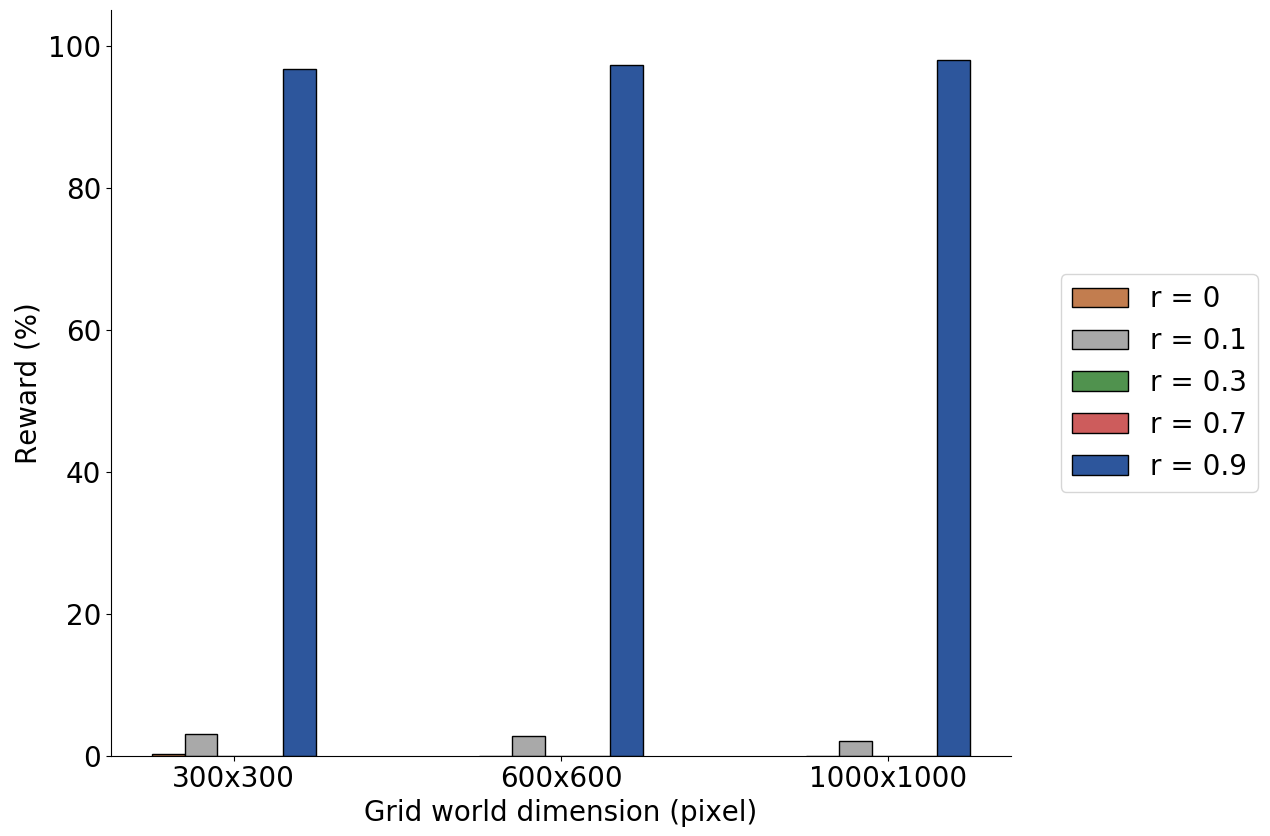}
\caption{Reward distribution of the RL agent on different grid resolutions over $1000$ episodes}
\label{fig:dc-captcha-results}
\end{figure}

In this section, we propose a divide and conquer technique to defeat the reCAPTCHA  system for any grid size without retraining the RL agent. The idea consists in dividing the grid into sub-grids of size $100\times100$ and then applying our trained agent on these sub-grids to find the optimal strategy for the bigger screen (see Figure ~\ref{fig:dc-captcha}). Figure~\ref{fig:dc-captcha-results} shows that this approach is effective and the success rates for the different tested sizes exceed $90\%$. 

\section{Effect of cell size}
\begin{figure}[h]
    \centering
    \subfloat[cell size 1x1 pixel]{{\includegraphics[width=5cm]{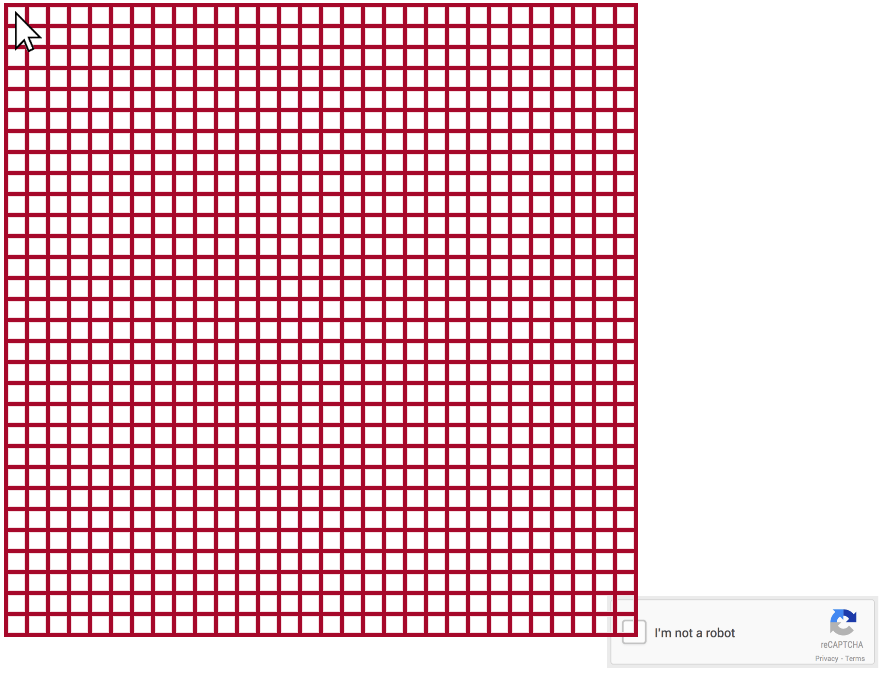} }}
    \qquad
    \subfloat[cell size 3x3 pixel]{{\includegraphics[width=5cm]{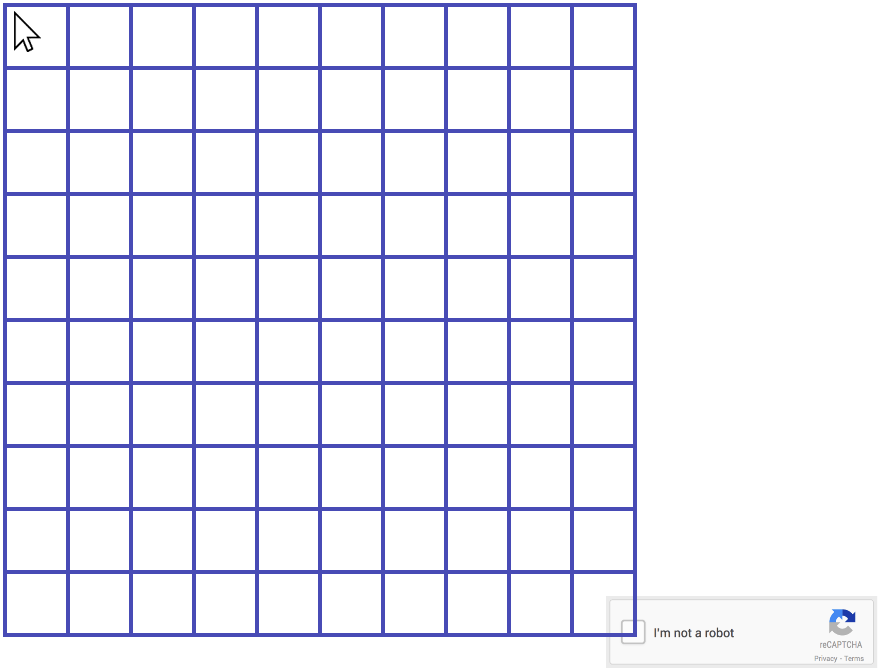} }}
    \caption{Illustration of the effect of the cell size on the state space}
    \label{fig:grid-cell-sizes}
\end{figure}

Here, we study the sensitivity of our approach to the cell size as illustrated in Figure~\ref{fig:grid-cell-sizes}.

\begin{figure}[h!]
\centering
\includegraphics[width=0.45\linewidth]{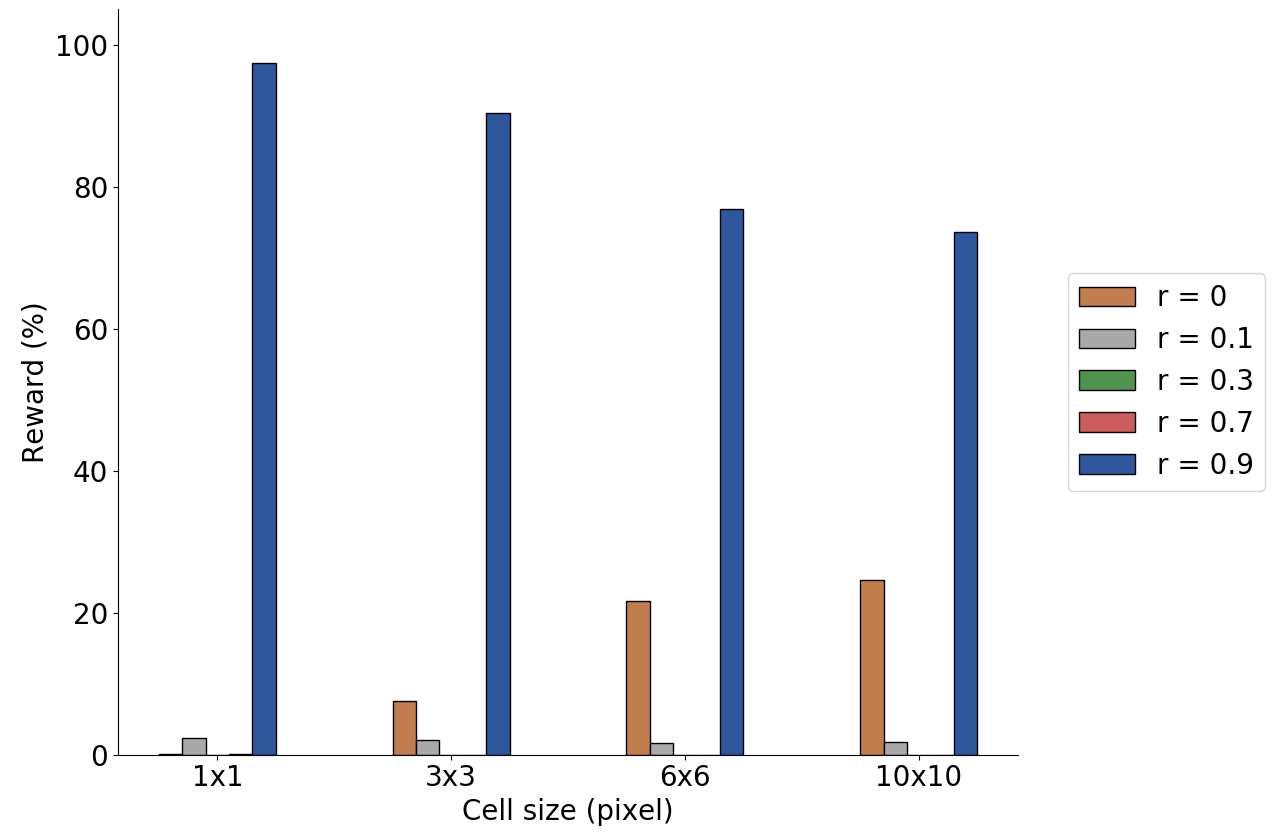}
\caption{Reward distribution for different cell sizes over $1000$ episodes}
\label{fig:cellsize-results}
\end{figure}

Figure~\ref{fig:cellsize-results} illustrates the obtained performance. We observe that when the cell size increases, the success rate of the agent decreases. For, cell size of $10$, the RL agent is detected as a bot in more than $20\%$ of the test runs. We believe that this decline is explained by the fact, with a big cell size, the agent scheme will contain more jumps which may be considered as non-human behavior by the reCAPTCHA  system.

\section{Details of the reCAPTCHA  environment}\label{sec:details-implementation}
Most previous works (e.g \cite{Bock2017}) used the browser automation software \textit{Selenium} \cite{selenium} to simulate interactions with the reCAPTCHA  system. At the beginning, we adopted the same approach but we observed that the reCAPTCHA  system always returned low scores suggesting that the browser was detected as fake. After investigating the headers of the HTTP queries, we found an automated header in the webdriver and some additional variables that are not defined in a normal browser, indicating that the browser is controlled by a script. This was confirmed when we observed that the reCAPTCHA  system with \textit{Selenium} and a human user always returns a low score.

It is possible to solve this problem in two different ways. The first consists in creating a proxy to remove the automated header while the second alternative is to launch a browser from the command line and control the mouse using dedicated Python packages such as the \textit{PyAutoGUI} library \cite{pyautogui}. We adopted the second option since we cannot control the mouse using \textit{Selenium}. Hence, unlike previous approches, our environment does not use browser automation tools.

Another attempt to use \textit{Tor} \cite{tor} to change the IP address did not pass the reCAPTCHA  test and resulted in low scores (i.e $0.3$). It is possible that the reCAPTCHA  system uses an API services such as \textit{ExoneraTor} \cite{exonerator} to determine if the IP address is part of the Tor network or not on a specific date.

We also discovered that simulations running on a browser with a connected Google account receive higher scores compared when no Google account is associated to the browser.

To summarize, in order to simulate a human-like experience, our reCAPTCHA  environment (1) does not use browser automation tools (2) is not connected using a proxy or VPN (3) is not logged in with a Google account.
\iffalse
\begin{itemize}
    \item does not use browser automation tools,
    \item is not connected using a proxy or VPN,
    \item is not logged in with a Google account
\end{itemize}
\fi
\section{Conclusion}
This paper proposes a RL formulation to successfully defeat the most recent version of Google's reCAPTCHA. The main idea consists in modeling the reCAPTCHA  test as finding an optimal path in a grid. We show how our approach achieves more than $90\%$ success rate on various resolutions using a divide and conquer strategy. This paper should be considered as the first attempt to pass the reCAPTCHA  test using RL techniques. Next, we will deploy our approach on multiple pages and verify if the reCAPTCHA adaptive risk analysis engine can detect the pattern of attacks more accurately by looking at the activities across different pages on the website.

\bibliographystyle{unsrt}
\bibliography{rldmsubmit}

\end{document}